\documentclass[utf8]{FrontiersinHarvard} 

\usepackage{url,hyperref,lineno,microtype,subcaption}
\usepackage[onehalfspacing]{setspace}
\usepackage{amsmath}
\usepackage{xfrac}
\usepackage{booktabs}
\usepackage{multirow}
\usepackage{bm}
\usepackage{caption}

\def\keyFont{\fontsize{8}{11}\helveticabold }
\def\firstAuthorLast{C. Galazis {et~al.}} 
\def\Authors{Christoforos Galazis\,$^{1,2,*}$, Ching-En Chiu\,$^{2,3}$, Tomoki Arichi\,$^{4}$, Anil A. Bharath\,$^{5}$ and Marta Varela\,$^{2}$}
\def\Address{$^{1}$Department of Computing, Imperial College London, London, UK \\
$^{2}$National Heart \& Lung Institute, Imperial College London, London, UK \\
$^{3}$Department of Electrical Engineering, Imperial College London, London, UK \\
$^{4}$Centre for the Developing Brain, King's College London, London, UK \\
$^{5}$Department of Bioengineering, Imperial College London, London, UK}
\def\corrAuthor{Christoforos Galazis}

\def\corrEmail{c.galazis20@imperial.ac.uk}

\begin{document}
\onecolumn
\firstpage{1}

\title[PINNing Cerebral Blood Flow]{PINNing Cerebral Blood Flow: Analysis of Perfusion MRI in Infants using Physics-Informed Neural Networks} 

\author[\firstAuthorLast ]{\Authors} 
\address{} 
\correspondance{} 

\extraAuth{}

\maketitle

\begin{abstract}

Arterial spin labeling (ASL) magnetic resonance imaging (MRI) enables cerebral perfusion measurement, which is crucial in detecting and managing neurological issues in infants born prematurely or after perinatal complications. However, cerebral blood flow (CBF) estimation in infants using ASL remains challenging due to the complex interplay of network physiology, involving dynamic interactions between cardiac output and cerebral perfusion, as well as issues with parameter uncertainty and data noise.
We propose a new spatial uncertainty-based physics-informed neural network (PINN), SUPINN, to estimate CBF and other parameters from infant ASL data. 
SUPINN employs a multi-branch architecture to concurrently estimate regional and global model parameters across multiple voxels. It computes regional spatial uncertainties to weigh the signal.  
SUPINN can reliably estimate CBF (relative error $-0.3 \pm 71.7$), bolus arrival time (AT) ($30.5 \pm 257.8$), and blood longitudinal relaxation time ($T_{1b}$) ($-4.4 \pm 28.9$), surpassing parameter estimates performed using least squares or standard PINNs. Furthermore, SUPINN produces physiologically plausible spatially smooth CBF and AT maps.
Our study demonstrates the successful modification of PINNs for accurate multi-parameter perfusion estimation from noisy and limited ASL data in infants. Frameworks like SUPINN have the potential to advance our understanding of the complex cardio-brain network physiology, aiding in the detection and management of diseases.
Source code is provided at: \url{https://github.com/cgalaz01/supinn}.

\tiny
 \keyFont{ \section{Keywords:} Physics-informed neural networks, Cardiac-brain network physiology, Neuroimaging, Arterial spin labelling, Cerebral blood perfusion} 
\end{abstract}

\section{Introduction}
 
Arterial spin labelling (ASL) is a non-invasive magnetic resonance imaging (MRI) technique that measures cerebral blood flow (CBF) without exogenous contrast agents \citep{lindner2023current}. CBF maps can be computed on a voxel-by-voxel basis by fitting mathematical models of haemodynamics based on ordinary differential equations (ODEs) \citep{alsop2015recommended}. These models help capture the complex temporal dynamics of blood flow, which is essential for understanding the intricate cardiac-brain network physiology. This understanding may aid diagnosing and managing various conditions, such as some forms of dementia and stroke \citep{rossi2022heart,tahsili2017heart}.

The bidirectional cardiac-brain network physiology operates as an intricate system where the heart and brain continuously influence each other \citep{candia2024measures}. The heart supplies oxygenated blood to the brain, affecting cerebral perfusion and pulsatile flow \citep{silverman2020physiology}, while the brain regulates cardiac function through the two autonomic nervous systems, the sympathetic and parasympathetic \citep{gordan2015autonomic}. This network incorporates feedback loops such as cerebral autoregulation and neurovascular coupling to maintain optimal function \citep{claassen2021regulation}.

In infants, particularly those with conditions like congenital heart disease (CHD) or preterm birth, this network is especially vulnerable due to immature autoregulation and developmental sensitivity \citep{de2024effects,claassen2021regulation}. These factors can result in altered cerebral haemodynamics, leading to issues such as delayed brain maturation, an increased risk of cerebral white matter injury, and potentially adverse long-term neurodevelopmental outcomes \citep{mcquillen2010effects}. Preterm neonates are often admitted to hospital to receive external physiological support whilst their bodies mature, which brain perfusion must be sufficient during this period.

The infant demographic thus benefits from non-invasive CBF monitoring techniques like ASL \citep{counsell2019fetal}. ASL can provide insights into the complex physiological interplay between the heart and brain, guiding interventions to support optimal brain development and overall cardiovascular healthy \citep{mcquillen2010effects,castle2022cardiac}.

A thorough understanding of this cardiac-brain network is crucial for managing infant health. Specifically, it is essential for optimizing neuroprotection strategies, improving surgical and medical management, and enhancing the long-term neurodevelopmental prospects of these infants \citep{de2023postnatal}. However, further research is needed to fully understand the independent effects and mechanisms of cardio-cerebral coupling \citep{castle2022cardiac,meng2015cardiac}, particularly in the developing infant brain \citep{baik2021cardiac}. Achieving this understanding in infants will require the development of even more accurate CBF monitoring techniques than those currently available.

Computing voxel-by-voxel CBF maps is achieved by fitting mathematical models of haemodynamics based on ODEs \citep{alsop2015recommended}. Many of these perfusion model ODEs assume very simplified physiology (e.g., plug blood flow to the brain, single magnetization compartments in the brain) and can therefore be solved analytically \citep{buxton1998general,alsop2015recommended}. It is often further assumed that the perfusion model parameters are perfectly known. In these conditions, CBF is estimated from a single perfusion-weighted image (PWI). These assumptions do not apply to CBF estimates in pathological conditions or groups with heterogeneous physiological properties, such as infants. 

Imaging infants, particularly those born preterm, presents further challenges due to lower signal-to-noise ratio (SNR). This is attributed to lower baseline CBF and longer arrival times (AT) of the magnetically labelled bolus \citep{dubois2021mri,varela2015cerebral}. Additionally, the need for higher spatial resolution in smaller infant brains further reduces SNR \citep{dubois2021mri}. Motion during scanning is also common in infants, further degrading image quality and leading to artifacts \citep{dubois2021mri,varela2015cerebral}.  

Unfortunately, voxel-by-voxel ASL analysis is susceptible to spatial inconsistencies, amplified by the lower SNR noise in infant perfusion weighted image (PWI) signals \citep{krishnapriyan2021characterizing,wang2022and}. Haemodynamic models are challenging to parameterise in the infant population due to dramatic physiological changes in the first weeks of life, during which most physiological parameters differ substantially from adult values. This is true of haemodynamic variables such as CBF, and also tissue composition, reflected in MR relaxation time constants such as $T_1$ and $T_2$. This is further complicated by the limited availability of data in this demographic \citep{de2023postnatal}.

In adult ASL, CBF estimation is commonly performed at a single time point following labelling \citep{detre2012applications}. This relies on several assumptions about haemodynamics and MR parameters that do not usually hold for infants. Given the complexity of the cerebral blood flow network in infants, past ASL studies in infants have therefore acquired PWIs at multiple time points following labelling to enable the simultaneous estimation of haemodynamic parameters beyond CBF, such as AT \citep{varela2015cerebral}. Past studies estimated CBF and other parameters using methods such as least squares fitting (LSF) using the analytical solution to the perfusion ODE \citep{varela2015cerebral}. However, due to the complexity of haemodynamic models, most model parameters need to be estimated separately. The lack of methods capable of simultaneously estimating both local and global parameters presents a significant challenge.

CBF has been estimated from infant ASL data using optimizers like LSF \citep{varela2015cerebral} and Bayesian estimation \citep{pinto2023modelling}, where adult models are fitted to the PWI signal. These voxel-by-voxel approaches often struggle with the very noisy PWIs typical of infant data, especially when estimating several parameters at once. Recently, neural network (NN)-based techniques for parameter estimation have become increasingly popular. NNs have demonstrated a remarkable ability to make accurate predictions even from noisy and corrupt data \citep{tian2020deep,hernandez2022recent}. However, such performance typically requires vast amounts of training data \citep{tian2020deep}, which is currently not available for infants \citep{korom2022dear,hernandez2022recent,de2023postnatal}.

Physics-informed neural networks (PINNs) \citep{karniadakis2021physics}, an emerging branch of machine learning, integrate physical laws (expressed as differential equations, DEs) into machine learning models. This approach improves a network's predictive capabilities even with limited and noisy data, as the DE agreement terms effectively act as a strong regularizer \citep{karniadakis2021physics}. PINNs can simultaneously solve DEs (forward problem) and estimate system parameters (inverse problem) from sparse experimental data. This makes them well-suited for biomedical applications \citep{ghalambaz2024physics}, evident by their increased usage in fields such cardiovascular \citep{moradi2023recent,herrero2022ep,sahli2020physics,van2022physics,kissas2020machine} and brain \citep{sarabian2022physics,kamali2023elasticity,de2023spatio,min2023non} research.

In cardiovascular studies, PINNs have been successfully applied to predict electrophysiological tissue properties from action potential recordings \citep{herrero2022ep} and to diagnose atrial fibrillation by estimating electrical activation maps \citep{sahli2020physics}. Additionally, PINNs have been used to quantify myocardial perfusion using MR imaging \citep{van2022physics} and to predict arterial pressure by analyzing MRI data of blood velocity and wall displacement \citep{kissas2020machine}. However, while PINNs are typically robust to noise, they suffer from the spatial inconsistencies associated with voxel-by-voxel fitting. PINNs' performance is notoriously variable, especially in inverse mode \citep{bajaj2023recipes}.

A significant challenge in PINN development is that they are often tested using synthetic data, which may not be a robust benchmark for performance on experimentally-acquired data. This is because few biomedical problems described by differential equations have known analytical solutions. Consequently, applications like CBF estimation using ASL data present rare opportunities to test PINNs' performance directly on experimental data and compare it to established parameter estimation methods such as LSF. Such real-world applications are crucial for validating and improving PINN methodologies in biomedical research.

This study introduces and evaluates PINNs as a tool for reliably estimating haemodynamic parameters from noisy infant ASL images. We propose a novel PINN framework, named Spatial Uncertainty PINN (SUPINN), which incorporates two key noise-mitigating improvements: 1) \textbf{Regional}: We assume neighbouring voxels share similar local parameters (e.g., CBF and AT) and therefore similar time courses. We thus propose weighting the confidence in each measurement by its spatial variability. 2) \textbf{Global}: For global parameters (e.g. $T_{1b}$), which are identical across all voxels within a subject, our multi-branch SUPINN learns from multiple voxels simultaneously to estimate a shared global parameter. Our method is particularly suited for imaging data acquired with limited and noisy samples over a given time period.

\section{Methods}
Our source code is publicly available at: \url{https://github.com/cgalaz01/supinn}.

\subsection{Dataset}
ASL brain MRI studies were conducted on seven infants aged 32 to 78 weeks postmenstrual age. The cohort included three infants with no pathology, one with periventricular leukomalacia, one with basal ganglia and white matter atrophy along with mild ventriculomegaly, one with agenesis of the corpus callosum, brain atrophy, and mild ventriculomegaly, and one with mild ventriculomegaly. Although this study does not include infants with known cardiac impairment, it is sufficient as our focus at this stage is on evaluating PINNs within the available diverse cohorot.

\begin{figure}[htbp!]
\centering
  \includegraphics[width=0.85\textwidth]{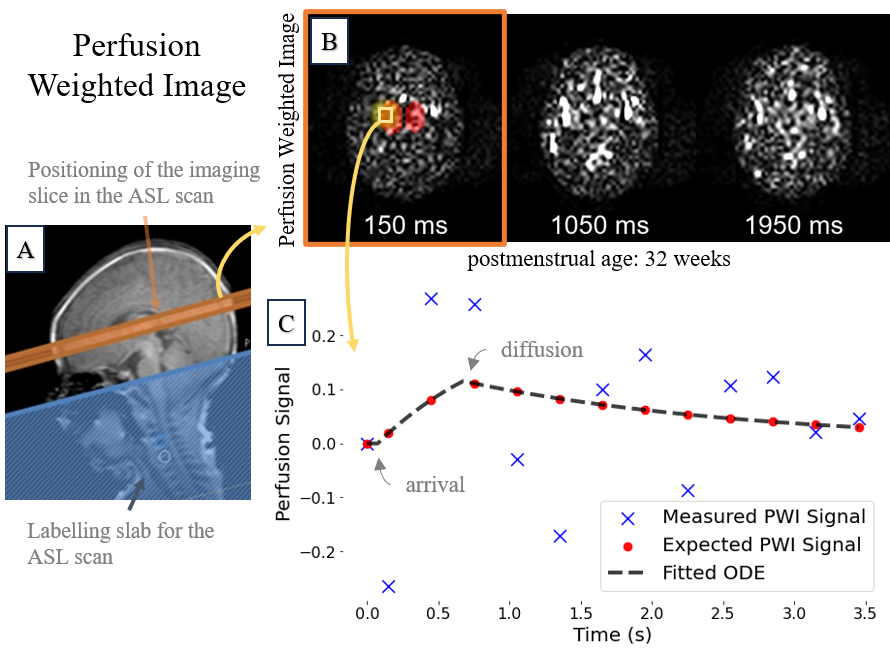}
  \caption{A representative 32-week postmenstrual case showing: A) $T_2$-weighted image highlighting the ASL imaging slice (orange); B) Subsampled perfusion-weighted image time series; and C) The measured perfusion signal of a single voxel over the entire duration, along with the corresponding ground-truth analytical model (see Eq. \ref{eqn:ode}).}
  \label{fig:pwi_example}
\end{figure}

All images were acquired in a Philips 3T Achieva scanner using an 8-element head coil under ethical approval following informed parental consent (REC: 09/H0707/83). PWIs were acquired on a single mid-brain transverse plane at 12 time points (every 300 ms) following a single pulsed labelling event \citep{petersen2006non}, at a spatial resolution of $3.04 \times 3.04 \times 5.5 ~mm^3$. For a representative PWI time series and accompanying signal plot, refer to Figure \ref{fig:pwi_example}.

In all subjects, our analysis focused on a manually segmented region of interest that includes the thalami and basal ganglia (Figure \ref{fig:pinn}). This deep grey matter region shows better SNR and fewer partial volume effects than cortical grey matter.

\subsection{Mathematical Model for ASL}

The relationship between the PWI signal, $S(t)$, and CBF can be expressed as the temporal convolution between an arterial input function, $AIF(t)$, and a tissue response function, $R(t)$: $S = AIF \ast R $ \citep{buxton1998general}. AIF is a top-hat function, here with a known duration $\tau = 900~ms$, that arrives at each voxel at a variable $t=AT$, and $R(t)$ is dominated by magnetization relaxation over venous outflow. As in \citep{alsop2015recommended}, we assume that the longitudinal magnetization relaxation of the blood is well described by $T_{1b}$ throughout.

We neglect the effect of the repeated excitation pulses on apparent $T_{1b}$ and assume that all PWI scaling constants are known, as in \citep{varela2015cerebral}. Then:
\begin{equation}
\label{eqn:asl_signal}
    S(t) = 
    \begin{cases}
        0 & \text{if $t < AT$} \\
        CBF \times (t - AT) \times e^{\sfrac{-t}{T_{1b}}} & \text{if $AT \leq t < AT + \tau$} \\
        CBF \times \tau \times e^{\sfrac{-t}{T_{1b}}} & \text{if $AT + \tau \leq t$}
    \end{cases}
\end{equation}

This model can be differentiated to yield an ODE defined in 3 branches:

\begin{equation}
\label{eqn:ode}
    \frac{dS}{dt} =
    \begin{cases}
        0 & \text{if $t < AT$} \\
        CBF \times e^{\sfrac{-t}{T_{1b}}} \times \left ( 1- \frac{t - AT}{T_{1b}} \right ) & \text{if $AT \leq t < AT + \tau$} \\
        -CBF \times e^{\sfrac{-t}{T_{1b}}} \times \frac{\tau}{T_{1b}} & \text{if $AT + \tau \leq t$}
    \end{cases}
\end{equation}

The three branches in Eqs. \ref{eqn:asl_signal} and \ref{eqn:ode} depict three distinct signal evolution phases: the periods before, during, and after the arrival of labelled blood at each voxel. We found that approximating the discontinuous 3-branched ODE in equation \ref{eqn:ode} using a NN leads to poor convergence properties. To circumvent this issue, we combine the three phases using smoothing hyperbolic tangent functions (see Supplementary Table S1).

\subsection{Ground Truth Estimation}
An auxiliary MRI scan was used to estimate ground-truth $T_{1b}$ in each subject \citep{varela2011method}. Then, a robust LSF was performed using the analytical haemodynamic model in Eq \ref{eqn:ode} to estimate ground-truth CBF and AT on a voxel-by-voxel basis.

Most biomedical problems described by DEs do not have an analytical solution and can only be solved numerically. For these, the accuracy of parameter identification methods is typically estimated using \textit{in silico} data, which do not capture the complexities of experimental measurements. The existence of an analytical ASL haemodynamic model (Eq \ref{eqn:ode}) presents a unique opportunity to test on experimental data the accuracy of model parameter estimation methods such as PINNs.

\subsection{Loss Function and Training Scheme}
PINNs are optimized to learn a solution that both matches the data and satisfies known known cardiac-brain network physiology principles. They minimize the combined loss function defined as: $ \mathcal{L} = \mathcal{L}_{ODE} + \gamma \mathcal{L}_{data}$. Due to the high noise in the data, $\mathcal{L}_{data}$ is weighted using an empirically-set coefficient $\gamma = 0.005$. Initial conditions, $S(t=0)=0$, are enforced by rescaling $S(t)$ using a hyperbolic tangent function \citep{lu2021deepxde}.

$\mathcal{L}_{ODE}$ measures the agreement with Eq. \ref{eqn:ode}. This loss is calculated by evaluating the residual of the differential equation at a set of collocation points ($N_O$) using the network's predictions and taking the mean squared error:
\begin{equation}
\label{eqn:loss_ode}
    \mathcal{L}_{ODE} = \frac{1}{N_O} \sum_i^{N_O} \left ( \frac{d\hat{s}}{dt}(t_i) - f(t_i, \hat{s}(t_i) \right )^2
\end{equation}

$\mathcal{L}_{data}$ is the data loss, which measures the mean squared error between the network's PWI estimation and the values measured across the 12 time points ($N_D$) acquired in each voxel:

\begin{equation}
\label{eqn:loss_data}
    \mathcal{L}_{data} = \frac{1}{N_D} \sum_i^{N_D} \left ( {w_{t_i} \times ||\hat S(t_i) - S(t_i)||}^2 \right ),
\end{equation}
where $w = 1$ is the weight of each PWI time point. $w$ is used in SUPINN with details available in section \ref{sec:supinn}.

When optimizing the PINNs' weights, we propose a three-tier hierarchical optimization scheme (see Supplementary Table S2). We initially optimize the PINNs in forward mode, focusing on aligning the network approximately with the underlying ODE without estimating specific parameters. We then solve the ODE in inverse mode to estimate the local parameters CBF and AT, and the global parameter $T_{1b}$. We finalise by fine-tuning the parameter estimation.

\subsection{PINN Architecture}   
\label{sec:pinns}
PINNs are implemented using DeepXDE v1.11 \citep{lu2021deepxde} and TensorFlow v2.15 \citep{abadi2016tensorflow}.
As a baseline PINN architecture \citep{raissi2019physics,karniadakis2021physics}, we use a fully connected neural network with hyperbolic tangent activation functions and two hidden layers, each consisting of 32 units. It includes one input unit for time $t$ and one output unit for the PWI signal $S(t)$.

\subsection{SUPINN Architecture}
\label{sec:supinn}

\begin{figure}[htbp!]
\centering
  \includegraphics[width=1.0\textwidth]{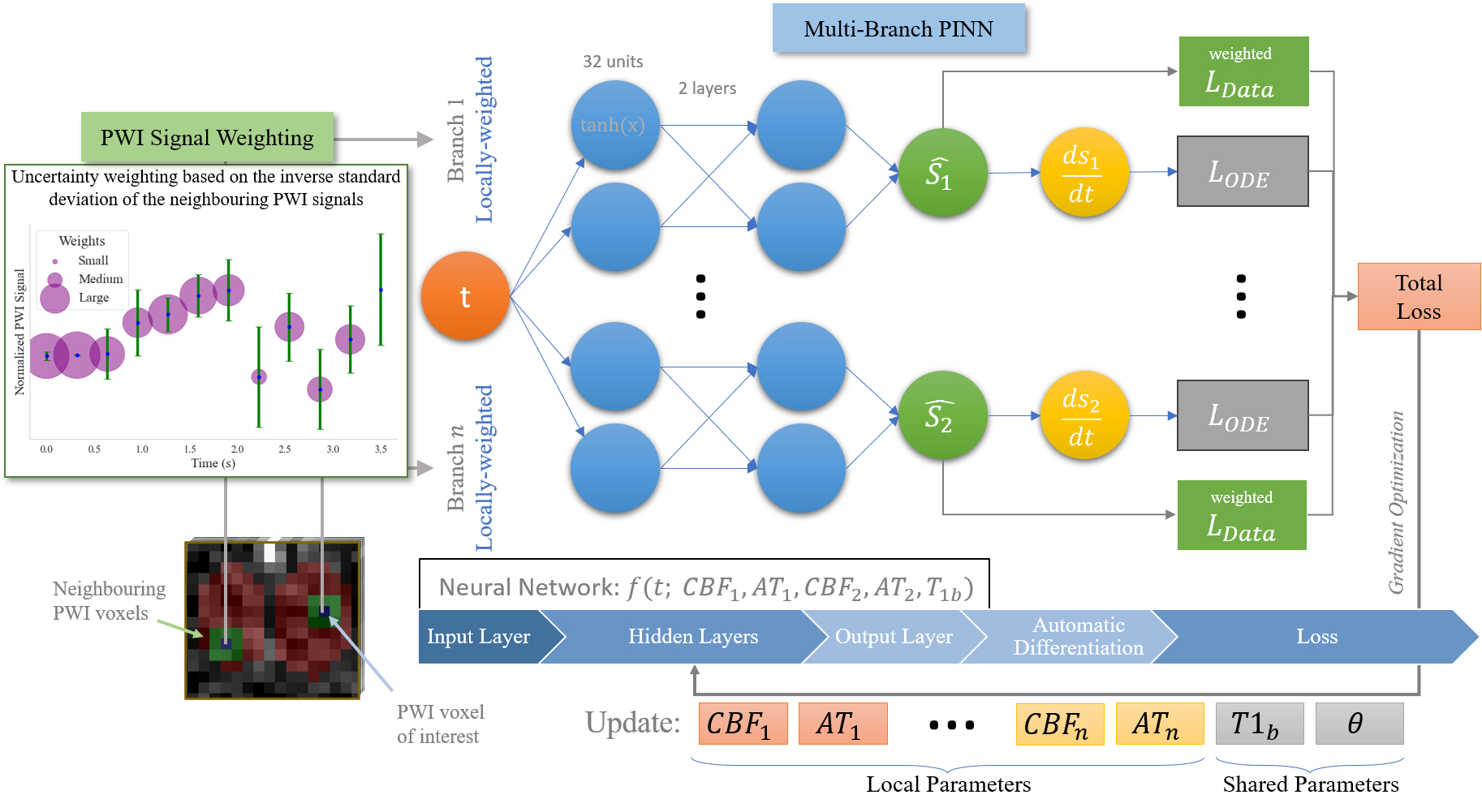}
  \caption{Overview of our proposed SUPINN model, depicted here in a 2-branch variant for illustration purposes, but adaptable to larger configurations. This study employs a 3-branch model based on empirical findings.
  }
  \label{fig:pinn}
\end{figure}

The baseline PINN models the signal from each voxel separately, ignoring the spatial relationships between the different sets of measurements. We expect, however, that neighbouring voxels have similar CBF and AT values, with deviations primarily due to noise. To incorporate this information in the model, we propose a spatial uncertainty PINN, SUPINN (Figure \ref{fig:pinn}). SUPINN inversely weighs the contribution of each PWI time point, $w$ (see Eq \ref{eqn:loss_data}), by their uncertainty levels. The uncertainty is estimated by calculating the standard deviation of the PWI signal in immediate neighbouring voxels within the region of interest at a given time point: $w_t = 1 / \sqrt{\frac{\sum{(S(t_i) - \mu_{t_i})^2}}{8}}$, where $w$ is the weight at time point $t$. The weights for each voxel across time are then scaled such that the highest uncertainty corresponds to a weight of $w=0.1$ and the smallest uncertainty to $w=1$. The weights in data loss $\mathcal{L}_{data}$ (Eq \ref{eqn:loss_data}) are updated accordingly.

SUPINN use a multi-branch architecture to reliably estimate global (subject-specific) parameters, such as $T_{1b}$ by pooling information from more than one voxel. It simultaneously estimates voxel-specific parameters $CBF$ and $AT$. The subnetworks' graphs are merged, allowing information sharing through backpropagation.

Each SUPINN branch employs the baseline PINN architecture described in section \ref{sec:pinns}. Voxels are selected randomly within the region of interest, with preliminary findings suggesting that a 3-branch network combining data from 3 voxels gives the optimal balance between estimation accuracy and computational efficiency. While voxel-specific CBF and AT parameters are estimated independently in each branch, $T_{1b}$ is shared across the selected voxels. The loss function, $\mathcal{L}$, for this architecture is the sum of the data agreement and ODE agreement losses (Eqs. \ref{eqn:loss_ode} and \ref{eqn:loss_data}) for each branch: $\mathcal{L} = \sum_i^{N=3} \mathcal{L}_{i, ODE} + \mathcal{L}_{i, data}$.

\subsection{Experimental Setup}
We compared SUPINN against several benchmarks: a standard PINN (Section \ref{sec:pinns}), a robust LSF method \citep{varela2015cerebral}, and a modified LSF (LSF-multi) that averages parameter estimations from 3 selected voxels. As we have limited data, evaluation against deep NN is not currently possible. All computations were performed on a 3XS Intel Core i7. The average execution times per voxel were approximately 0.05s for LSF/LSF-multi, 31s for PINN, and 40s for SUPINN.

Evaluation metrics include the mean and standard deviation of the relative error (RE), computed as $(predicted - target) / target \times 100$ for each parameter. When a method led to CBF estimates that increasingly diverged from ground truth CBF by more than one order of magnitude after $50K$ iterations, it was deemed not to have converged. These failed estimates were not taken into account when assessing the quantitative performance of each method. We compute a method's convergence rate as $|total - failed| / total \times 100$. The spatial smoothness of CBF and AT was assessed using the mean and standard deviation of the Laplacian variance across subjects \citep{pertuz2013analysis}, where lower variance signifies greater spatial parameter homogeneity. We also estimate the mean squared error (MSE) between the prediction and ground truth PWI signal (forward mode).

\section{Results}

Our proposed SUPINN architecture, designed to address variable data noise levels and simultaneously estimate local and global parameters, showed excellent performance on infant ASL data (see Table \ref{tab:results}). SUPINN showed improvements in both PWI signal (forward) and parameter (inverse) estimations compared to the standard PINN and LSF/LSF-multi methods at the cost of increased computational time.

SUPINN led to more accurate parameter estimates, especially for CBF. Specifically, SUPINN achieved a RE of $-0.3 \pm 71.7$ for CBF, $30.5 \pm 257.8$ for AT, and $-4.4 \pm 28.9$ for $T_{1b}$. Additionally, the predicted PWI signal closely matched the ground truth, as evidenced by the smallest MSE of $0.4 \pm 0.8$, as shown in Table \ref{tab:results}. Finally, both the base PINN and SUPINN achieved high parameter convergence rate, with rates of $99.9\%$ and $100\%$, respectively.

\begin{table}[htbp!]
\centering
\caption{Summary of the convergence rate, relative error and Laplacian variance for CBF, AT and $T_{1b}$, and mean squared error of the predicted solution. A model's quality is indicated by a low standard deviation and a mean error close to 0.}
\label{tab:results}
\resizebox{\textwidth}{!}{%
\begin{tabular}{@{}lccccccc@{}}
\toprule
\multicolumn{1}{c}{\multirow{2}{*}{Model}} & \multirow{2}{*}{\begin{tabular}[c]{@{}c@{}}Convergence\\ Rate (\%)\end{tabular}} & \multicolumn{3}{c}{Relative Error (\%)}                                                                     & \multicolumn{2}{c}{Laplacian Variance}            & Mean Squared Error \\
\multicolumn{1}{c}{}                       &                                   & CBF                                      & AT                              & $T_{1b}$                       & CBF                          & AT                 & PWI Signal ($\times10^{-3}$)                       \\ \midrule
LSF                                        & 62.6                              & $390.7 \pm 1306.7$ \medspace   & $53.8 \pm 510.7$ \medspace      & $-43.1 \pm 32.2$ \medspace     & $29.1 \pm 11.8$ \medspace    & $3.1 \pm 2.7$      & $26.9 \pm 22.7$                   \\
LSF-multi                                  & 96.4                              & $549.7 \pm 1272.0$ \medspace   & $121.9 \pm 467.0 $\medspace     & $-31.4 \pm 29.9$ \medspace     & $12.4 \pm 5.7$ \medspace     & $1.2 \pm 1.0$      & $38.3 \pm 31.4$                   \\
PINN                                       & 99.9                              & $96.0 \pm 475.8$ \medspace     & $68.6 \pm 283.9$ \medspace      & $8.6 \pm 35.9$ \medspace       & $0.5 \pm 0.4$ \medspace      & $0.5 \pm 0.8$      & $1.1 \pm 1.3$                     \\
SUPINN                                     & \bm{$100.0$}                        & \bm{$-0.3 \pm 71.7$} \medspace & \bm{$30.5 \pm 257.8$} \medspace & \bm{$-4.4 \pm 28.9$} \medspace & \bm{$0.4 \pm 0.4$} \medspace & \bm{$0.1 \pm 0.1$} & \bm{$0.7 \pm 0.8$}                     \\ \bottomrule
\end{tabular}
}
\end{table}

We typically observe higher noise levels in the PWI signal of younger infants. Despite this challenge, Supplementary Figure S1 shows that SUPINN consistently achieved lower RE in CBF across all subjects compared to other methods despite low SNR. Additionally, SUPINN achieved the most accurate estimates of AT and $T_{1b}$ in the majority of cases. Notably, SUPINN also demonstrated resilience in estimating parameters for infants with neurological disorders (indicated with an asterisk in the figure).

The robustness of our model is further demonstrated in Supplementary Figure S2, where we evaluated its performance on synthetic signals. White Gaussian noise was added to each of the synthetically generated PWI signals, with the standard deviation progressively increased in increments of 0.1, up to a maximum of 0.5. Despite increasing the standard deviation of the noise, SUPINN maintained stable parameter estimations, especially for CBF and AT. This highlights the model's ability to handle noisy data effectively. In comparison, the baseline PINN also exhibited resilience in estimating AT and $T_{1b}$, but its CBF estimations deteriorated progressively as the noise level increased. On the other hand, the LSF method showed the greatest sensitivity to noise, with parameter estimations degrading noticeably even with a small amount of added noise.

\begin{figure}[htbp!]
\centering
  \includegraphics[width=1.0\textwidth]{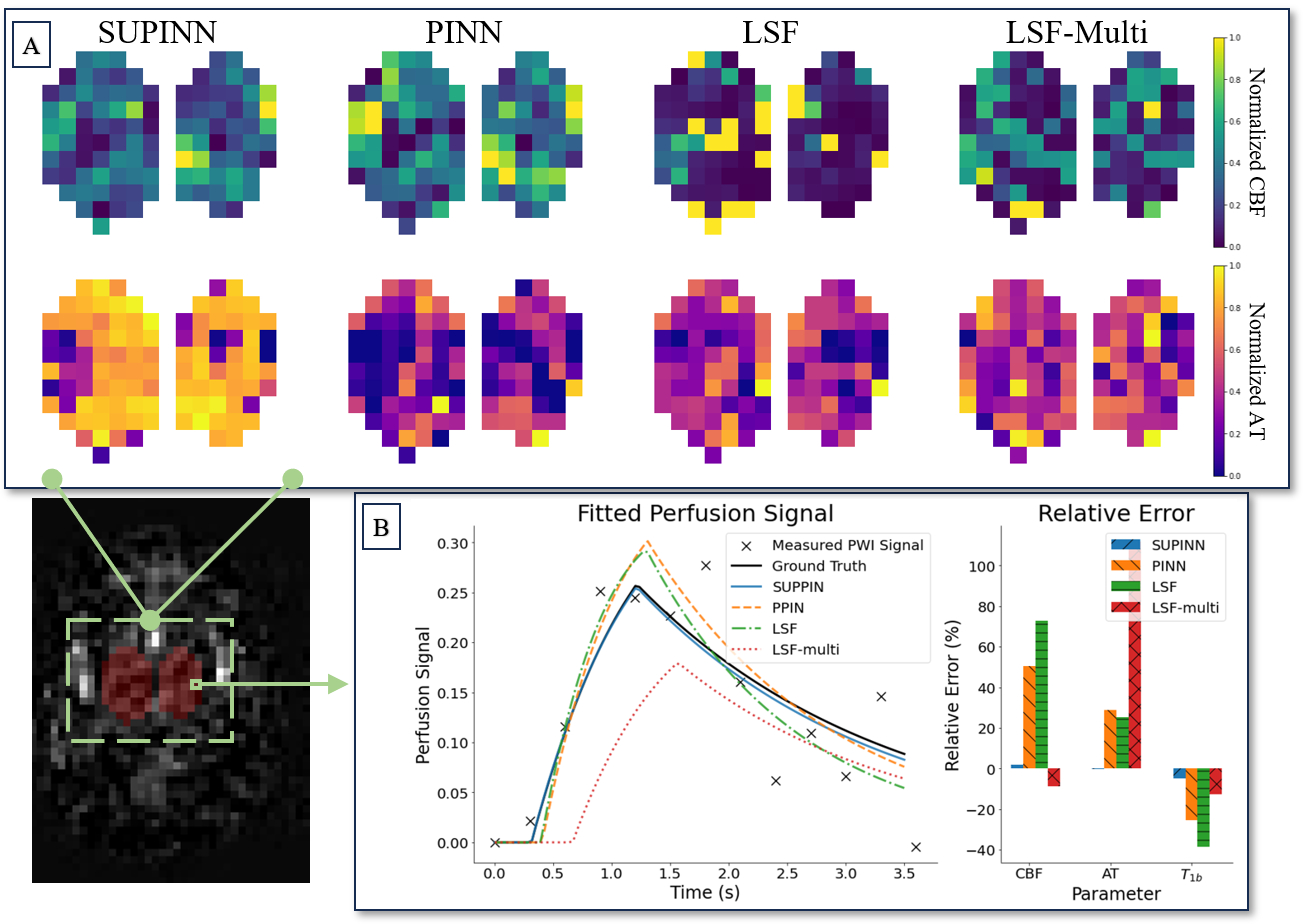}
  \caption{Panel A shows spatial maps of parameter estimation in deep grey matter for a subject aged 32 weeks. Each row corresponds to the normalized relative error of two parameters: CBF and AT. The columns display the estimation results from four methods: SUPINN, PINN, LSF, and LSF-multi. Panel B depicts the parameter relative error of the models for a single voxel.}
  \label{fig:maps}
\end{figure}

Figure \ref{fig:maps} illustrates the spatial maps of the CBF and AT predictions for a representative infant. The SUPINN estimates, shown in the first column, exhibit higher spatial consistency for both CBF and AT compared to other methods. This consistency is quantified by the lowest Laplacian variance achieved, as detailed in Table \ref{tab:results}). Specifically, SUPINN attained a Laplacian variance of $0.4 \pm 0.4$ for CBF and $0.1 \pm 0.1$ for AT across all cases, indicating smoother and more reliable spatial predictions.

\section{Discussion}
We introduce SUPINN, a novel multi-branch PINN technique for estimating parameters from noisy data. By solving ODEs over neighbouring regions with similar properties and estimating uncertainty through voxel comparisons, SUPINN simultaneously estimates local and global parameters with high accuracy. We test it on the challenging task of estimating haemodynamic parameters from extremely noisy infant multi-delay ASL data, where it outperforms both standard PINNs and LSF.

SUPINN's strong performance is also underpinned by our three-tier optimization regime, use of hard initial conditions and the replacement of non-differentiable transitions in the baseline model (Eq. \ref{eqn:ode}) by a smoothly interpolated version. These enhancements are crucial for accurately capturing the complex cerebral haemodynamics in infants, in whom subtle alterations in perfusion can have implications for brain development.

LSF is widely used for parameter identification from various medical images, including ASL. It performs reliably when estimating a small number of parameters, particularly multiplicative factors or temporal intervals (such as CBF or AT in Eq. \ref{eqn:ode}). Following the literature \citep{varela2015cerebral,hernandez2022recent}, we used robust LSF to estimate ground-truth CBF and $AT$ when separate ground-truth measurements of $T_{1b}$ were available. LSF is nevertheless extremely unreliable when estimating exponents such as $T_{1b}$ in conjunction with CBF and AT.

PINNs have several advantages over LSF other than improved overall performance. Evidently from SUPINN, they offer a framework for more flexibly combining data from different brain and, in the future, cardiac regions. Contrary to standard PINNs, SUPINN is able to handle data with high noise to further improve performance. SUPINN lead to spatially smoother CBF and AT maps within the same brain region, aligning more closely with physiological expectations. Additionally, PINNs can be applied to ODEs with no known analytical solutions, opening up the possibility of using more sophisticated and personalized perfusion ODEs.

The multi-delay ASL data are well-suited for testing parameter identification methods, as the existence of an analytical solution allows for easy application of LSF. SUPINN's performance can therefore be directly evaluated on real, noisy clinical data. This is in contrast to most PINN studies, which are typically evaluated on synthetic data with known noise distributions. Although our current dataset does not include cases of CHD in infants, the techniques developed here are likely to be applicable to such cases, given the similar challenges in analyzing cerebral haemodynamics.

Future work will expand the evaluation to include a larger infant cohort of both healthy and CHD cases to validate the robustness and generalizability of SUPINN. This will enable us to assess the efficacy of the improved CBF estimation specifically in the context of CHD and explore its relation to the disease. Optimizing voxel selection strategies and exploring alternative PINN architectures, such as graph-based approaches, can further improve performance by better representing spatial relationships critical in various clinical scenarios, including CHD.

SUPINN's applicability extends to other problems where ODEs are solved over neighbouring regions with similar parameters. SUPINN can, for example, contribute to estimating quantitative MRI properties (such as $T_1$ or $T_2$) by simultaneously solving the Bloch equations in neighbouring voxels within the same tissue \citep{zimmermann2024pinqi}.

This paper proposes SUPINN, a PINN method able to handle noisy data by leveraging spatial information. We demonstrate its potential to improve the characterisation of haemodynamics using infant ASL. With further refinement and validation, SUPINN can become a valuable clinical tool, providing precise and accurate physiological data for diagnosis, monitoring, and treatment planning in various clinical contexts, including potential applications in infants with CHD.

\section*{Conflict of Interest Statement}
The authors declare that the research was conducted in the absence of any commercial or financial relationships that could be construed as a potential conflict of interest.

\section*{Author Contributions}
The author contributions are as following: CG - Software, Methodology, Writing – original draft; CC - Software, Writing – review \& editing; TA - Data curation, Writing – review \& editing; AB - Supervision, Writing – review \& editing; MV - Conceptualization, Supervision, Data curation, Writing – review \& editing.

\section*{Funding}
This work was supported by the UK Research and Innovation (UKRI) Centres of Doctoral Training (CDT) in Artificial Intelligence for Healthcare (AI4H) \url{http://ai4health.io} (Grant No. EP/S023283/1), the NIHR Imperial Biomedical Research Centre (BRC), and the British Heart Foundation Centre of Research Excellence at Imperial College London (RE/18/4/34215).

\section*{Acknowledgments}
We acknowledge computational resources and support provided by the Imperial College Research Computing Service (\url{http://doi.org/10.14469/hpc/2232}).

\bibliographystyle{Frontiers-Harvard}
\bibliography{main.bib}


\end{document}